\IfFileExists{scrartcl.cls}{\documentclass{scrartcl}}{\documentclass{article}}
\usepackage[top=2cm, bottom=2.25cm, left=2.5cm, right=2.5cm]{geometry}
\usepackage[T1]{fontenc}
\usepackage[utf8]{inputenc}
\usepackage{amsmath, amsthm, amssymb}
\IfFileExists{dsfont.sty}{\usepackage{dsfont}}{}
\usepackage{graphicx}
\usepackage{color}
\IfFileExists{subcaption.sty}{\usepackage{subcaption}}{}
\usepackage{caption} \usepackage{subcaption} 
\IfFileExists{multicol.sty}{\usepackage{multicol}}{}
\IfFileExists{bm.sty}{\usepackage{bm}}{}
\IfFileExists{dlfltxbcodetips.sty}{\usepackage{dlfltxbcodetips}}{}
\IfFileExists{tabularx.sty}{\usepackage{tabularx}}{}
\IfFileExists{csquotes.sty}{\usepackage{csquotes}}{}
\usepackage{url}

\renewcommand{\phi}{\varphi}

\numberwithin{equation}{section}

\theoremstyle{definition} 
\theoremstyle{definition} 
\theoremstyle{definition} 

\title{Accelerating nanodrug development in continuous flow systems using informed prediction models based on low-cost surrogate nanoparticles}

\author{
	Kai Dahms\textsuperscript{1,*}, Eilien Heinrich\textsuperscript{1,*}, Jochen Schmid\textsuperscript{2,*},\\
	Michael Bortz\textsuperscript{2}, Iryna Savych\textsuperscript{1}, Regina Bleul\textsuperscript{1}\\ 
	\small \textsuperscript{1}Fraunhofer Institute for Microengineering and Microsystems (IMM), 55129 Mainz, Germany\\ \small \textsuperscript{2}Fraunhofer Institute for Industrial Mathematics (ITWM), 67663 Kaiserslautern, Germany\\ 
	\small jochen.schmid@itwm.fraunhofer.de
}

\date{}

\begin{document}
\maketitle

\begingroup 
\renewcommand{\thefootnote}{\fnsymbol{footnote}} \footnotetext[1]{These authors contributed equally to this work.} \endgroup

\begin{abstract}
\small{\noindent
The development of nanotherapeutics often involves extensive empirical optimization due to the sensitivity of nanoparticle properties, such as size and polydispersity index (PDI), to minor changes in process parameters. Factors like formulation concentration, flow rates, and mixing ratios can significantly influence clinical efficacy and therapeutic outcomes. The absence of predictive mathematical frameworks has made iterative experimental screening necessary, increasing both costs and development time.

This study introduces and validates a predictive modeling approach based on shape constraints, aiming to enhance the estimation of nanoparticle characteristics across various process conditions. Using controlled microfluidic methods, liposomes and lipid nanoparticles were systematically prepared under varying lipid concentrations, flow rates, and aqueous-to-organic mixing ratios.

The shape-constrained model, informed by both experimental data and expert knowledge, was subsequently validated for a pharmaceutical application using minimal empirical data. Results reveal that shape-constrained modeling facilitates accurate prediction of nanoparticle size and dispersity, reducing the need for extensive experimental workflows. This framework supports rational and efficient process development for manufacturing nanomedicine systems.
}
\end{abstract}

\section{Introduction}
\label{sec:introduction}

Nanoparticle-based delivery systems have become an increasingly important platform in modern pharmaceutical development, offering new opportunities to overcome limitations associated with conventional drug formulations. By enabling the protection, transport, and controlled release of sensitive or challenging therapeutic molecules, nanocarriers have opened new avenues for the treatment of complex diseases, including cancer and severe infectious diseases as well as the prevention as vaccines. Depending on their composition and intended application, nanocarrier systems comprise a broad range of platforms, including lipid-based systems such as liposomes and lipid nanoparticles (LNPs), polymeric carriers, inorganic nanoparticles, and biogenic delivery systems. These technologies can improve key pharmaceutical properties such as drug solubility, stability, biodistribution, and release kinetics, while supporting targeted delivery strategies and potentially reducing systemic side effects \cite{ref1}.

The clinical relevance of these concepts is demonstrated by several nanomedicine products that have already reached the market. Liposomal formulations such as Caelyx\textregistered{} (pegylated liposomal doxorubicin) and AmBisome\textregistered{} (liposomal amphotericin B) represent early examples of how nanoparticle-based approaches can modify the pharmacological behavior of established drugs. By altering drug distribution and improving tolerability, these formulations have contributed to reduced adverse effects, such as cardiotoxicity or nephrotoxicity, while maintaining therapeutic efficacy \cite{ref2,ref3}. These examples illustrate how nanocarriers have evolved from a promising research concept into clinically validated drug delivery platforms.

More recently, lipid nanoparticles have gained global attention through their pivotal role in mRNA-based vaccines. The rapid development and successful deployment of LNP-based vaccines such as Comirnaty\textregistered{} demonstrated the potential of nanocarriers to enable entirely new therapeutic modalities and accelerated their transition from specialized drug delivery systems to a key technology platform in modern medicine.

\begin{figure}[htbp]
\centering
\includegraphics[width=\textwidth]{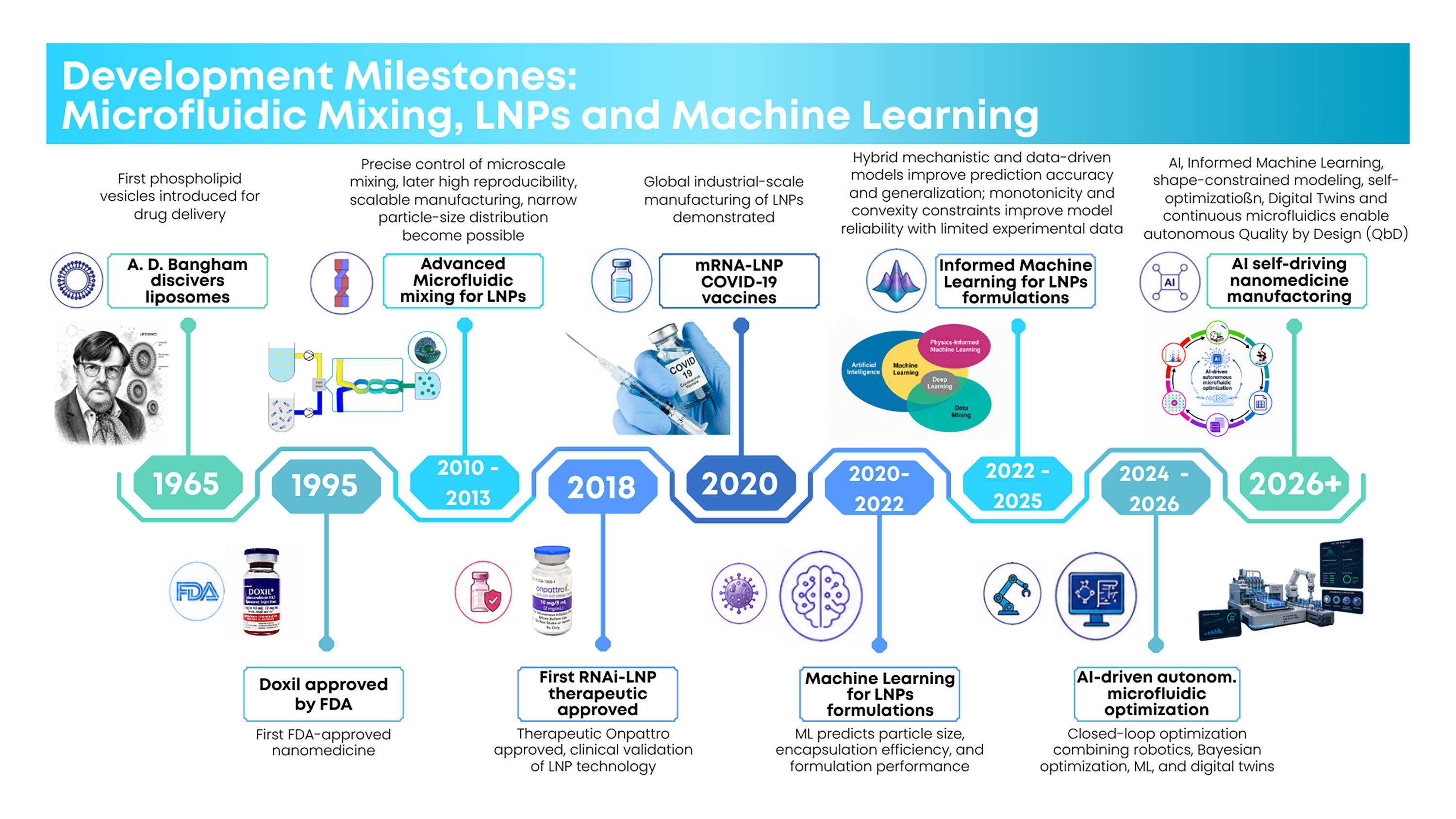}
\caption{Development Milestones: Microfluidic Mixing, LNPs and Machine Learning.}
\label{fig:development-milestones}
\end{figure}

The evolution of nanocarrier-based delivery systems is the result of decades of scientific progress across multiple disciplines. As illustrated in Figure~\ref{fig:development-milestones}, key development milestones span from the first description of liposomes in the 1960s to the emergence of microfluidic manufacturing platforms and the clinical translation of LNP-based mRNA therapeutics in the 2020s \cite{ref4,ref5}. Early foundational work established the physicochemical principles of lipid-based self-assembly, while subsequent advances in polymer science and lipid chemistry expanded the range of available carrier platforms. The introduction of microfluidic mixing technologies marked a pivotal step towards reproducible and scalable nanoparticle production, enabling precise control over particle size and encapsulation \cite{ref6}. Most recently, the successful deployment of mRNA-LNP vaccines against COVID-19 demonstrated the clinical maturity of these systems and catalyzed a new wave of research into next-generation nanocarrier platforms, including ionizable lipid systems, targeted delivery strategies, and AI-assisted formulation development \cite{ref7}. This timeline underscores a clear trend: the field has progressively shifted from empirical discovery towards rational, data-driven design. This transition forms the central motivation of the present work.

However, the increasing complexity of nanocarrier-based products also highlights significant challenges in their development and manufacturing. Unlike conventional formulations, where relatively small process variations may have limited impact, nanoparticle systems can be highly sensitive to changes in production parameters. Variations in lipid composition, mixing conditions, or flow rates can significantly influence critical quality attributes such as particle size and polydispersity index (PDI) \cite{ref8}. These parameters directly affect biological performance, including pharmacokinetics, biodistribution, and therapeutic efficacy \cite{ref5}. Consequently, nanocarrier manufacturing requires highly controlled processes and often relies on iterative formulation optimization to achieve reproducible product quality.

The systematic Quality by Design (QbD) approach further enhances manufacturing reliability, aligning nanoparticle formulation processes with regulatory guidelines like those issued by the FDA \cite{ref9}. QbD defines the Quality Target Product Profile (QTPP), which specifies desired product characteristics and connects them to Critical Quality Attributes (CQAs), Critical Material Attributes (CMAs), and Critical Process Parameters (CPPs) \cite{ref10,ref11,ref12}.

For mRNA-LNP systems, key CQAs include particle size, PDI, RNA encapsulation efficiency, lipid ratios, and zeta potential \cite{ref13}. CMAs such as lipid composition and N/P ratios, alongside CPPs like flow rates and mixing conditions during microfluidic production, influence these attributes significantly. Optimizing CQAs through deliberate control of CMAs and CPPs is therefore essential to achieve scalable production with high efficacy and safety \cite{ref10,ref13}.

The traditional empirical approach to designing nanocarriers often involves extensive experimentation and trial-and-error processes, which increase time, resource consumption, and costs. Design-of-experiments (DoE) methods provide a structured framework to evaluate key manufacturing parameters and their interactions, while mapping operational design spaces for scalable nanoparticle production \cite{ref10}. Central to this process is ensuring robust quality control, encompassing critical evaluations of particle size, polydispersity index (PDI), morphology, encapsulation efficiency, and drug release profiles. Such quality measures are pivotal for translating laboratory-scale processes into reproducible and scalable pharmaceutical production.

Recent innovations in machine learning (ML) and artificial intelligence (AI) have accelerated the design and optimization of nanoparticle systems \cite{ref14}. Supervised ML methods facilitate the prediction of nanoparticle properties, such as transfection efficiency and morphology, using extensive datasets of lipid compositions and processing parameters \cite{ref15}. For example, Bayesian optimization and ensemble modeling techniques have significantly improved precision in tuning lipid formulations and optimizing microfluidic conditions for mRNA-LNP preparation \cite{ref16}. Additionally, AI-driven systems, including digital twins and real-time monitoring capabilities, enhance production through predictive modeling \cite{ref17} and automated adjustments, streamlining resource-intensive experimentation and maximizing operational efficiency \cite{ref18,ref19}.

Further advancements involve integrating process analytical technologies (PAT) and real-time release testing (RTRT) to improve manufacturing control, predictive maintenance, and monitoring \cite{ref20}. These tools facilitate quality assurance and minimize variability, supporting scalable production while reducing costs. Nevertheless, key challenges persist, particularly concerning the selection of novel excipients under intellectual property (IP) constraints and regulatory requirements \cite{ref21}. Helper lipids and PEGylated lipids also present manufacturing complexities that must be addressed to fully capitalize on AI and ML-driven methods for nanoparticle development.

In order to address these challenges, a combined data- and knowledge-based development strategy is established to improve the understanding and optimization of nanocarrier formulations. Specifically, informed shape-constrained modeling is applied to continuous microfluidic systems resulting in physics-informed, data-efficient optimization, improved process robustness, and reliable QbD implementation for nanodrug manufacturing. Moreover, this approach significantly reduces development costs by minimizing the number of required experiments and by leveraging cost-effective model formulations for training predictive models that can be transferred to expensive active pharmaceutical ingredients.

This article highlights the interplay between nanotechnology, predictive modeling, and AI in shaping the next generation of nanomedicines. By bridging empirical science with computational power, these approaches promise to reduce inefficiencies, streamline processes, and make innovative treatments accessible to patients worldwide.

\section{A combined data- and knowledge-based approach for nanocarrier development}
\label{sec:methodology}

In this paper, we propose an informed machine-learning approach \cite{ref22} combining experimental data and expert knowledge to improve the understanding and optimization of nanocarrier formulations. This approach not only helps identify suitable formulation conditions but also yields a structured framework that links formulation parameters, process conditions, and resulting nanoparticle characteristics. In particular, the proposed approach enables a more comprehensive understanding of the nanocarrier design space and supports the development of robust and reproducible manufacturing processes.

To validate the optimization model described in this article, the optimization of naproxen-loaded liposomes was employed as a pharmaceutical case study. Naproxen is a non-steroidal anti-inflammatory drug (NSAID) characterized by poor aqueous solubility (BCS Class II), which severely limits its oral bioavailability and therapeutic efficacy \cite{ref23,ref24}. Encapsulation within liposomal carrier systems represents a promising formulation strategy to overcome these biopharmaceutical limitations. The incorporation of naproxen into the lipid bilayer of liposomes enables enhanced solubilization, controlled drug release, and improved bioavailability at the target site \cite{ref25}.

\subsection{Experimental methodology and expert knowledge}
\label{subsec:experimental-methodology}

\begin{figure}[htbp] 
	\centering 
	\begin{subfigure}[t]{0.48\textwidth} \centering \includegraphics[width=\linewidth]{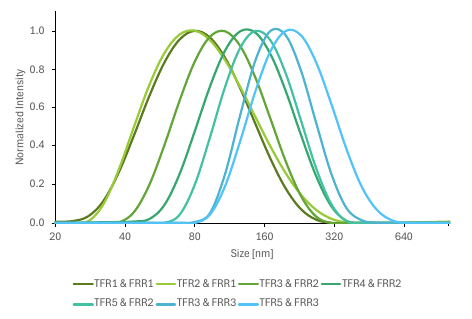} 
	\end{subfigure} 
	\hfill 
	\begin{subfigure}[t]{0.48\textwidth} \centering \includegraphics[width=\linewidth]{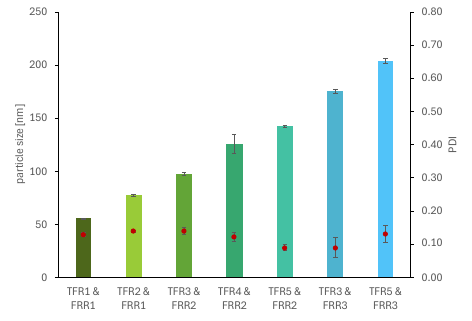} 
	\end{subfigure} 
	\caption{(a) Size deviation function of 7 different parameter combinations (TFR: Total Flow Rate; FRR: Flow Rate Ratio) of surrogate LNPs resulting in Z-averages between 50 nm and 200 nm. (b) Size (Z-averages) and PDI for 7 different parameter combinations. Size between 50 nm and 200 nm, PDI consistently less than 0.2.} 
	\label{fig:monotonicity-expert-knowledge} 
\end{figure}

Nanoparticle formulations were developed using microfluidic mixing techniques to ensure control and reproducibility. Microfluidic nanodrug formulation enables a precise tailoring of the characteristics such as size and PDI and encapsulation efficiency by rapidly mixing lipids and aqueous phase under laminar flow. Increasing the total flow rate enhances mixing efficiency by reducing the diffusion path length, while increasing the flow rate ratio of organic and aqueous phase dilutes the lipid phase more rapidly. Both effects generally result in smaller and more uniform nanoparticles \cite{ref6,ref23}. This monotonicity expert knowledge is illustrated in Figure~\ref{fig:monotonicity-expert-knowledge}.

To further facilitate cost-effective analysis, simplified nanotransporter systems were modelled (surrogate particles). For this purpose, relatively inexpensive analogous lipid substances resembling liposomes were used. These nanotransporters were subjected to different flow ratios
\begin{equation}
 r_{mix}:=\frac{F_{aq}}{F_{org}}
\end{equation}
of aqueous to organic medium and different total flow rates
\begin{equation}
 F_{tot}:=F_{aq}+F_{org}
\end{equation}
in a micromixer, thereby generating particles with varying hydrodynamic diameters and polydispersity indices.

The surrogate liposomes and surrogate lipid nanoparticles were produced using the caterpillar micromixer developed in-house at Fraunhofer IMM. This mixer has already shown to produce nanoformulations with constant product quality \cite{ref26} and assesses the ability to be used in a scaled up production \cite{ref27}. In-line dilution with filtered water regulated the ethanol concentration and ensured particle stability. The particle size distribution was subsequently analyzed using dynamic light scattering (DLS). Accordingly, all particle sizes reported in this paper are Z-averages.

\subsection{Informed particle size prediction}
\label{subsec:informed-particle-size-prediction}

In essence, the proposed approach for informed particle size prediction combines experimental data for the surrogate particles and a single data point for the target particle on the one hand with the aforementioned monotonicity expert knowledge on the other hand. It results in an informed size prediction model for pharmaceutically relevant particles. As it combines experimental data with expert knowledge, it allows for robust prediction at low experimental cost. Additionally, it allows for an informed and robust optimization of formulation parameters. In detail, the proposed particle-size prediction methodology proceeds in three steps, which are explained in the following three subsections.

\subsubsection{Shape-informed size prediction for surrogate particles}
\label{subsubsec:surrogate-size-prediction}

\begin{figure}[htbp]
\centering
\includegraphics[width=\textwidth]{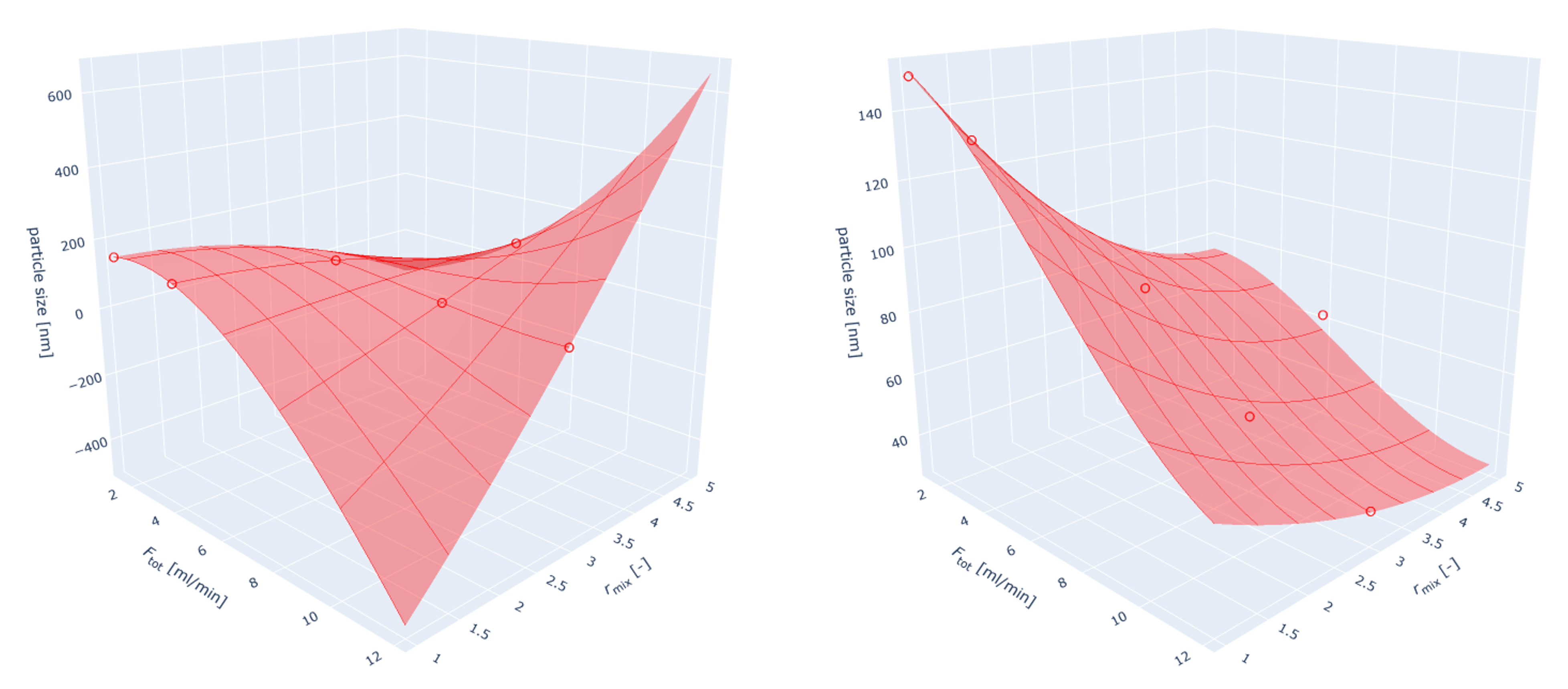}
\caption{(a) Unconstrained size prediction model. (b) Shape-constrained size prediction model. The unconstrained model glaringly violates process expert knowledge: It predicts negative particle sizes. Additionally, it is monotonically increasing with $r_{mix}$ for high total flow rates (right edge of the plot) and monotonically increasing with $F_{tot}$ for high flow rate ratios (rear edge of the plot). In contrast, the shape-constrained model is mathematically guaranteed to comply with the specified expert knowledge. The models were trained on the same 6 data points and both models are polynomial with the same polynomial degrees (degree 3 in the input $F_{tot}$ and degree 2 in the input $r_{mix}$).}
\label{fig:unconstrained-vs-shape-constrained}
\end{figure}

In the first step, an informed prediction model for the size of the chosen surrogate particle is trained, based on a small to moderate amount of experimental data for the surrogate particle and on the expert knowledge that the particle size decreases both with the total flow rate and the flow rate ratio. Specifically, this informed model is trained using the methodology of shape-constrained regression \cite{ref28,ref29} or, more specifically, monotonic polynomial regression with additional lower and upper bound constraints. An important feature of the shape-constrained regression algorithms developed in \cite{ref28,ref29} is that they produce models which are mathematically guaranteed to comply with the imposed shape expert knowledge. As a consequence, these models yield more reliable predictions and can extrapolate better than purely data-based regression models. Additionally, these models avoid overfitting, which purely data-based models often exhibit in the small-data regime. Figure~\ref{fig:unconstrained-vs-shape-constrained} compactly illustrates these advantages of shape-constrained regression models over standard unconstrained regression models. See also \cite{ref28,ref30,ref31,ref32} for applications of shape-constrained regression to other engineering domains.

\subsubsection{Shape-informed size prediction for pharmaceutically relevant particles}
\label{subsubsec:target-size-prediction}

\begin{figure}[htbp]
\centering
\includegraphics[width=\textwidth]{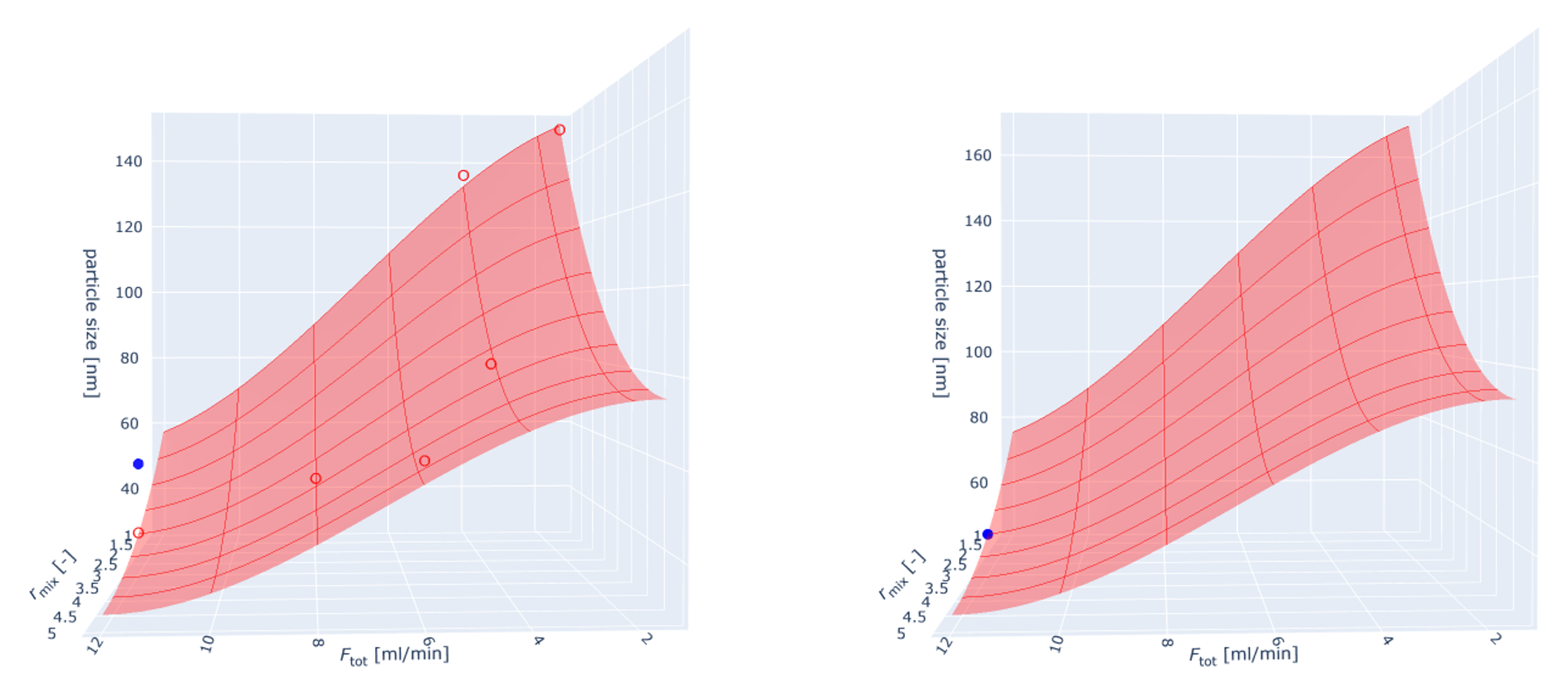}
\caption{(a) Shape-constrained size prediction model trained on the surrogate-particle data (red points). (b) Shape-constrained size prediction model shifted upwards to the target-particle anchor point (blue point).}
\label{fig:model-transfer}
\end{figure}

In the second step, the shape-informed prediction model from the first step is transferred from the surrogate particle to the considered pharmaceutically relevant particle. All that is necessary for this is one single data point for the target particle. It acts as an anchor point for the model transfer. Specifically, the model from the first step (Figure~\ref{fig:model-transfer}a) is shifted up- or downwards such that it passes exactly through this anchor point (Figure~\ref{fig:model-transfer}b). If more target particle data are available than just one, more sophisticated model transfer strategies can be applied than the simple vertical transport just described (and they will be explored in an upcoming paper). In the present paper, though, we confine ourselves to the simple vertical transport requiring only one target particle data point.

\subsubsection{Application for optimal formulation parameter choice}
\label{subsubsec:parameter-choice}

\begin{figure}[htbp]
\centering
\includegraphics[width=\textwidth]{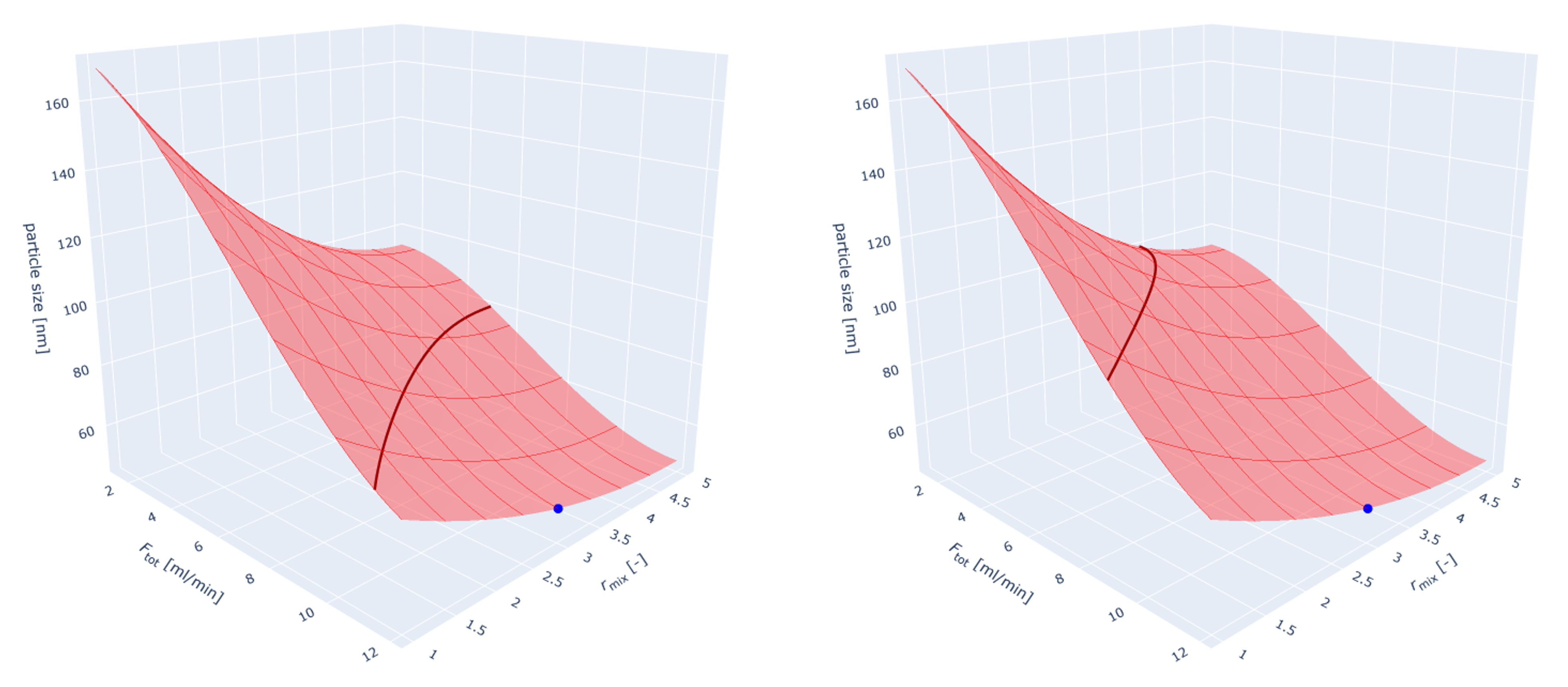}
\caption{Set of formulation parameter values for which the shape-informed prediction model predicts the target size of (a) 80 nm or of (b) 100 nm.}
\label{fig:target-size-contours}
\end{figure}

In the third step, the shape-informed prediction model from the second step is used to find optimal formulation parameters to produce pharmaceutically relevant particles with a desired target size. Specifically, the model is used to compute the set of all formulation parameter values for which the desired target size $y^*$ is to be expected. In Figure~\ref{fig:target-size-contours}, this set is depicted as the red contour line. And from this set, one can then select parameter settings one deems most appropriate for production. In this selection process, one can take into account further selection criteria like PDI minimization, for example. Since the model is imperfect, of course, the actual size of the produced particle will generally deviate from the desired target size value. As will be demonstrated below, however, the deviation is usually moderate. If it is not, the new data can be used to iteratively refine the model as indicated at the end of Section~\ref{subsubsec:target-size-prediction}.

\subsection{Case study with naproxen liposomes}
\label{subsec:case-study}

\begin{figure}[htbp]
\centering
\includegraphics[width=\textwidth]{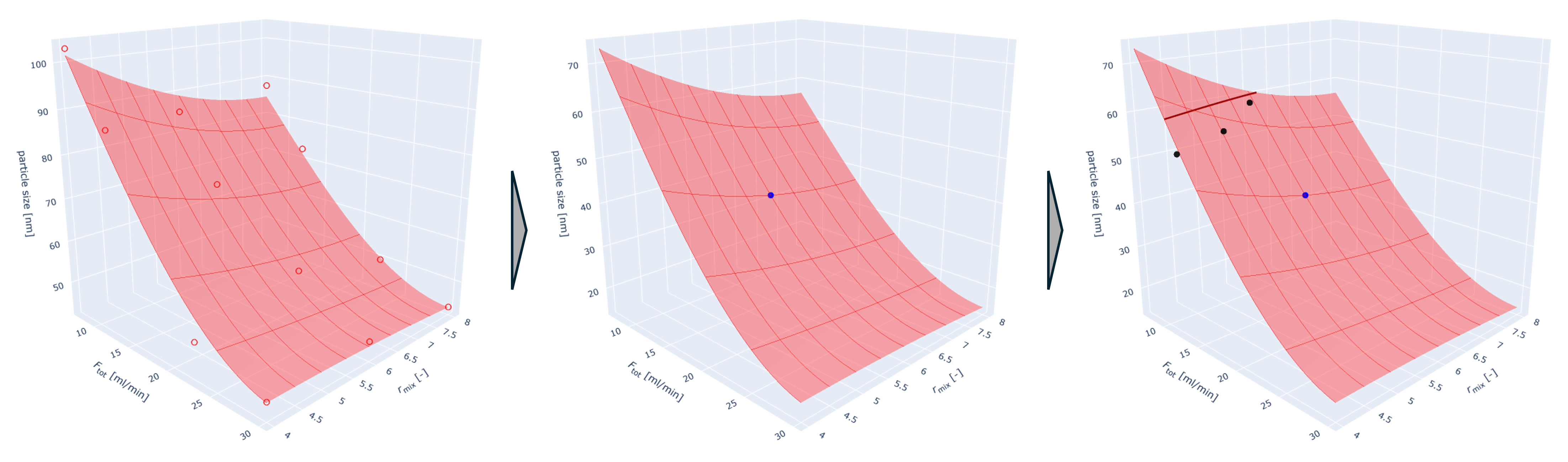}
\caption{Steps of the proposed informed particle size prediction methodology applied to naproxen liposomes. (a) Train shape-constrained model on 12 surrogate-particle data. (b) Shift that model downwards to the target-particle anchor point. (c) Use the shifted model to predict formulation parameters for target size 60 nm.}
\label{fig:naproxen-methodology}
\end{figure}

As a proof of concept, the general methodology described above was applied to naproxen liposomes. As the corresponding surrogate particle, liposomes made from a sorbitan fatty acid ester were produced. Naproxen liposomes were used, consisting of lecithin (S100 lipid), encapsulating a poorly water-soluble API (NSAID) within the lipid bilayer. The parameter space was defined with the total flow rate ranging from 3.75 to 30 mL/min and the flow rate ratio ranging from 4 to 8 in the microfluidic mixer. To avoid particle size variations caused by differing ethanol contents, the ethanol concentration was kept constant at 8\% by applying inline dilution in a subsequent T-piece mixing step. As a first step, 12 screening experiments were performed with the surrogate particle, and a shape-constrained size prediction model was trained based on those 12 experimental data points (Figure~\ref{fig:naproxen-methodology}a). As a second step, a single reference experiment was conducted with naproxen liposomes at the center of the parameter space (total flow rate: 15 mL/min, flow rate ratio: 6), yielding a particle size of approximately 40 nm. Using this single data point as an anchor, the shape-constrained model developed on the surrogate nanoparticle system was transferred (Figure~\ref{fig:naproxen-methodology}b) to predict process parameter combinations for the target formulation. Specifically, a target particle size of 60 nm was set, the corresponding 60 nm contour line was computed by the model, and three distinct process parameter combinations were then selected from that contour line by the experimenter (Figure~\ref{fig:naproxen-methodology}c).

\begin{figure}[htbp]
\centering
\includegraphics[width=0.7\textwidth]{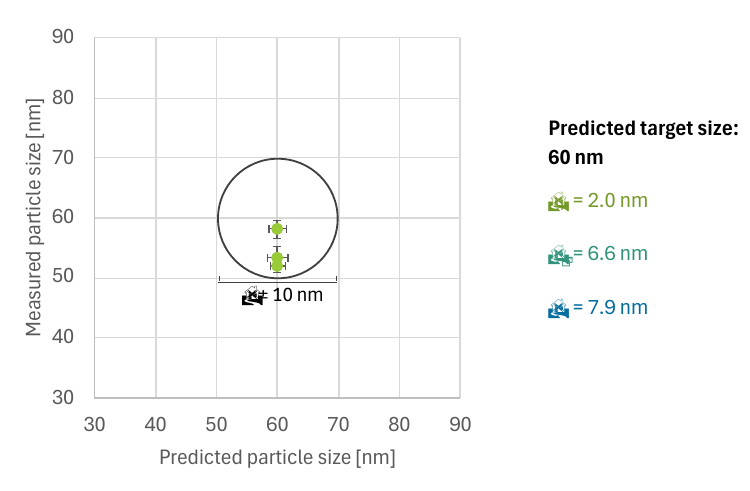}
\caption{Predicted versus measured particle size for 3 different parameter combinations targeting 60 nm naproxen liposomes.}
\label{fig:predicted-versus-measured}
\end{figure}

As can be seen in Figure~\ref{fig:predicted-versus-measured}, the measured particle sizes of all three predicted process parameter combinations fall within a range of $60\pm10$ nm, closely matching the defined target. Given the limited number of surrogate nanoparticle data points used for model training and only a single anchor measurement for the therapeutically active formulation, this represents a remarkably accurate and promising result, underscoring the predictive power and efficiency of the proposed approach.

\section{Conclusion}
\label{sec:conclusion}

The development of nanocarrier-based drug delivery systems requires a careful balance between formulation complexity, process understanding, and product performance. The approach presented in this work addresses this challenge by integrating systematic formulation development with data-driven strategies, thereby enhancing process understanding and supporting more efficient optimization workflows.

A key finding of this study is that cheap analog chemical structures can be employed to formulate surrogate nanoparticles. These serve as practical surrogates for therapeutically active formulations whose components, such as ionizable lipids, APIs, RNA, or DNA, are often considerably more expensive. By optimizing process parameters on the surrogate system first, material costs during formulation development can be substantially reduced. These surrogate particles enable the application of shape-constrained modeling to predict optimal process parameters for the target nanoparticle system. The relatively accurate prediction of particle size achieved with a limited dataset underlines the robustness and transferability of this approach. This suggests that additional process parameters, such as component concentration or temperature can be readily incorporated to further extend its predictive scope.

Apart from particle size, additional quality attributes of high therapeutic relevance, including encapsulation efficiency and transfection efficiency, represent promising targets for future model-based optimization. By linking process parameters, material characteristics, and resulting nanoparticle properties, the proposed framework offers a resource-efficient and versatile tool for the rational development of lipid-based nanocarrier formulations, with a clear potential for facilitating scale-up and reducing experimental burden.

\section*{Acknowledgments}
The authors gratefully acknowledge financial support from the Ministry of Science and Health (Rhineland-Palatinate, Germany) through the project SmartForm (724-0022\#2024/0005-1501 15402).

\begin{small}

\end{small}

\end{document}